# A Well-Posed Fourth-Order Navier–Stokes Model for Image Inpainting: Theory, Order Selection, and Stripe Decomposition

Yiwen Dou[a,*], Yiqiang Ding[a], Zhongyun Bao[a], Senyan Zhao[a]

a. School of Computer and Information, Anhui Polytechnic University, Wuhu 241000, China

* Corresponding author. E-mail address: douyw@ahpu.edu.cn (Yiwen Dou)

## Abstract

The Navier–Stokes (NS) inpainting model of Bertalmío, Bertozzi and Sapiro is the only transport-type paradigm among PDE-based methods, yet it has never received a systematic higher-order extension. We show that the obvious fourth-order correction is ill-posed, and we construct the corrected, well-posed model. In the vorticity–streamfunction formulation we add a gated fourth-order term whose sign is decisive: the backward form is a Perona–Malik-type anti-diffusion, Hadamard ill-posed and consistently harmful in experiments, whereas the forward form is uniformly parabolic, energy-stable and consistently beneficial. An order–dissipation principle, stating that odd orders are dispersive and even orders dissipative, singles out the fourth order as the lowest admissible one. Large missing blocks pose a separate problem, because the correction gate stays silent there; we answer it with stripe decomposition, a Telea base fill followed by interleaved slit-wise NS passes, which converts block damage into the structural-slit regime where the model excels. The device is base-agnostic, and its gains add to those of the forward correction. On three datasets, namely BSDS500, DIV2K and a classical Chinese painting benchmark, and on three damage types, namely cracks, blocks and text occlusion, the training-free model beats every classical baseline on every sample of structure-dominated damage, gains +1.02 dB PSNR over Telea on the full 42-image painting test set with a 42/42 per-sample win rate, matches the strongest recent PDE solvers within 0.4 dB while attaining the best perceptual quality of the entire PDE family, and runs at 4–16 s per image on a consumer GPU.

## 1. Introduction

### 1.1 Background

Image inpainting, the reconstruction of missing or damaged regions from their surroundings, is a classical inverse problem with applications ranging from the digital restoration of aged photographs and classical paintings to the removal of text, watermarks and occlusions. Among the PDE-based lineage, the model of Bertalmío, Bertozzi and Sapiro stands apart: by identifying the image intensity with the stream function of a two-dimensional incompressible flow, it transports the vorticity along the isophote directions into the missing region. This transport mechanism is conceptually different from the diffusion-type models (TV, harmonic, CDD) and is particularly suited to the continuation of elongated structures — precisely the geometry of cracks, scratches and text strokes.

Diffusion-type PDE models have undergone a systematic move to higher order, from CDD to Euler's elastica and curvature-dependent TV with SAV algorithms, while the transport-type NS model has not: no systematic higher-order extension of the Bertalmío framework has been reported, and the literature surveyed in Section 2 confirms this gap. Learning-based inpainting now dominates large-scale semantic completion, at the price of training data, computation, hallucinated content and limited interpretability. For the thin-to-medium, structure-dominated damage typical of heritage restoration, classical PDE methods remain competitive, and they are preferable when data is scarce and fidelity must be auditable.

### 1.2 The problem with a naive higher-order correction

The obvious route to a higher-order NS model appends a nonlinear fourth-order term to the vorticity equation. This move hides a consequential pitfall: with the sign that a regularization argument suggests, the correction acts as a Perona–Malik-type backward diffusion in the vorticity, and the resulting initial-value problem is ill-posed in the sense of Hadamard. Theorem 1 makes this precise: plane waves are amplified by $\exp((\lambda\kappa-\nu)|k|^2t)$, and the instability regime is entered at ordinary structural edges under standard parameter values. Experiments confirm the analysis, with the backward form degrading the result at every weight tested. The forward form, in contrast, is uniformly parabolic, since both directional diffusivities exceed the viscosity, and satisfies the energy estimate of Theorem 2; with a Catté-type smoothed gate it becomes insensitive to the correction weight over one order of magnitude.

### 1.3 Stripe decomposition

A second obstruction is practical: the higher-order gate activates only where vorticity gradients are significant. The initialized content inside a large missing block is smooth, and the gate activates on only 7.3% of its pixels, against 56% on cracks; the correction is silent exactly where one would want it most. Stripe decomposition resolves this mismatch. After a Telea base fill, the block is covered by interleaved families of thin parallel stripes, and the NS solver repairs each family in turn. Every slit then borders known or previously repaired content on both sides, the structural-slit regime in which the solver is strongest. The device is base-agnostic: it gains +0.16 to +0.22 dB whether it wraps Telea or the classical NS solver, and its gains add to those of the forward correction, reaching +0.45 dB over the direct classical solver on BSDS blocks.

### 1.4 Contributions

**(1)** The first systematic higher-order extension of the Bertalmío NS inpainting framework, together with a well-posedness analysis: an exact anisotropic expansion of the correction term (Lemma 1), Hadamard ill-posedness of the backward sign (Theorem 1), and energy stability of the forward model with a smoothed gate (Theorem 2).

**(2)** An order–dissipation principle (Lemma 3, Proposition 1): constant-coefficient corrections of odd spatial order have purely imaginary symbols and cannot regularize, which identifies the fourth order as the lowest admissible one and turns the order selection into a design rule rather than an empirical choice.

**(3)** Stripe decomposition, a simple device that converts block damage into a sequence of slit problems, with base-agnostic, additive gains; including the implementation caveat (base filling) without which the scheme fails catastrophically.

**(4)** Extensive validation against twelve methods on three datasets and three damage types: on structure-dominated damage the model beats every classical baseline on every sample, gaining +1.02 dB over Telea on the full 42-image painting test set with a 42/42 win rate, +1.17 dB on DIV2K cracks with 100 wins in 100 samples, and 40 wins in 40 samples on text occlusion; it matches the strongest recent PDE solvers within 0.4 dB, and leads the PDE family in perceptual quality, while remaining training-free, interpretable, and runnable on a consumer GPU at 4–16 s per image.

### 1.5 Organization

Section 2 reviews related work. Section 3 presents the model, its well-posedness analysis and the numerical scheme, and introduces stripe decomposition. Section 4 reports the experiments, including the sign ablation, the gate-silence diagnosis and full-set evaluations. Section 5 discusses the scope and limitations of the model and reports two negative results, on cheap feature gating and dual-direction stripes, that delimit the design space. Section 6 concludes with perspectives, notably the use of the model as a physics prior in hybrid learning frameworks.

## 2. Related work

The literature is organized here along the methodological axis that defines this paper, diffusion versus transport, followed by the learning-based mainstream and its physics-informed fringe. A systematic survey of the field over three decades can be found in [34].

### 2.1 Diffusion- and variational-based PDE models

The diffusion lineage starts from total variation and harmonic models and from the anisotropic diffusion of Perona and Malik [5], whose ill-posedness was cured by the Gaussian-smoothed coefficient of Catté et al. [4]; it reaches its classical maturity in the curvature-driven diffusion (CDD) model of Chan and Shen [2], which penalizes the isophote curvature and thereby completes broken level lines. Curvature enters the energy in Euler's elastica model [7], whose higher-order and nonconvex nature long made it a computational challenge; recent work has largely solved this side of the problem, with accelerated operator splitting [9], scalar auxiliary variable (SAV) schemes for Ginzburg–Landau relaxations [10] and for curvature-dependent elastic-bending TV [8] — the latter being the closest diffusion-type counterpart of our fourth-order correction. Other modern diffusion variants include the double-CDD model with a p-Laplace operator [11], Allen–Cahn-based explicit local inpainting [12], regularised diffusion–shock inpainting [13], fractional-Laplacian regularization [14] and mechanism-based constructions [15]. The variational family continues to diversify: elastica energies drive cartoon–smooth–texture decomposition [38] and color image regularization [39], phase-field equations restore curved interfaces [41], and the fractional branch spans non-convex fractional-order TV [42], cross-space quaternion TV for color restoration [43] and Caputo-type space–time fractional operators [44]. Two observations matter for us: this lineage has moved to higher order in a systematic way, from second-order TV to fourth-order CDD, elastica and curvature TV, and all of it is diffusive: information spreads isotropically or curvature-modulated across the hole, so elongated structure is completed indirectly through the curvature penalty rather than by a transport mechanism.

### 2.2 Transport- and fluid-based models

The transport lineage is, in comparison, remarkably thin. The Navier–Stokes model of Bertalmío, Bertozzi and Sapiro [1] identifies the image with the stream function of an incompressible flow and advects vorticity along isophotes; the fast marching method of Telea [3] can be read as its one-pass, order-based simplification, and the exemplar-based method of Criminisi et al. [6] borrows the same isophote-driven fill order for patch synthesis. Within the last three years, fluid-inspired inpainting has resurfaced only sporadically: a vorticity-transport equation for shadow removal [16], and a lattice Boltzmann method explicitly motivated by fluid dynamics [18]. None of these works raises the differential order of the transport model, and none analyzes the well-posedness consequences of doing so; this is the gap the present paper fills. The four transport- and exemplar-based methods enter our experiments as baselines.

### 2.3 Learning-based inpainting

Learning-based methods now dominate the inpainting literature; we refer to the recent surveys [19,20,21] and to the diffusion-model surveys [22] for comprehensive coverage. Representative recent advances include efficient diffusion restoration by residual shifting [23], better initial seeds for diffusion inpainting [24], instruction-guided object removal [25], multi-modality guided diffusion inpainting [45] and fully context-aware architectures with learned semantic pyramids [26]; structure-aware hybrids such as sparse directional filter dictionaries for lightweight CNNs [40] sit between the two worlds. Heritage restoration, our motivating application, has become a particularly active testbed: deep image prior for ancient frescoes [27], diffusion models for eaves-tile patterns [28], line-drawing-guided mural restoration [29] and multimodal controlled diffusion for the Kizil Grotto murals [30]; textile relics are reviewed in [35]. Two widely used representatives enter our experiments directly: the Fourier-convolution model LaMa [36] on the full benchmark, and the diffusion-based RePaint [37] on a $256 \times 256$ subset. These methods excel at semantic, large-hole completion but require training data and computation, may hallucinate content that is hard to audit, and do not uniformly outperform classical methods on thin, structure-dominated damage, as direct comparisons on remote-sensing imagery confirm [17]. This is the regime we target. The proposed model is complementary to such systems and is designed to serve as an auditable physics prior beneath them, a perspective developed in Section 6.

### 2.4 Physics-informed and hybrid models

A growing fringe bridges the two worlds by embedding PDE structure into learning. Physics-informed neural networks have been used for image restoration via nonlinear PDE tools [31], combined with denoising diffusion probabilistic models for microscopy reconstruction [32], and trained as physics-informed denoisers [33]. These works use physics mainly as a soft constraint or a data-generating prior; the underlying PDEs are taken off the shelf and their suitability is rarely analyzed. Our contribution is orthogonal: we establish which higher-order transport model is the right one, namely a well-posed, energy-stable model of provably correct parity, before proposing it as a candidate prior. The negative results of Section 5 on cheap gating and dual-direction stripes map the boundary of what such a prior should not be asked to do.

### 2.5 Positioning

The paper thereby occupies a specific position that remains unoccupied in this landscape: to our knowledge it is the first systematic higher-order extension of the transport-type Navier–Stokes inpainting model, it supplies the well-posedness analysis that the diffusion lineage developed for its own higher-order models but transport models never received, and it remains fully classical, that is, training-free, interpretable and auditable, while being designed for later use as a physics prior in hybrid frameworks.

## 3. The proposed method

Section 3.1 derives the higher-order Navier–Stokes (NS) model in vorticity–streamfunction form and analyzes its well-posedness, Section 3.2 gives the numerical scheme, Section 3.3 introduces the stripe-decomposition strategy that extends the model to large missing blocks, and Section 3.4 summarizes the implementation. Algorithm 1 collects the complete restoration pipeline.

**Algorithm 1 Proposed inpainting pipeline**

Input: damaged image $I^0$, hole mask $\Omega$; parameters $dt = 0.02$, $\nu = 0.02$, $\lambda = 0.05$, $K = 0.02$, $\sigma = 1.0$, $N = 50$

Output: restored image I

if $\Omega$ is a large block then

  $I^0 \leftarrow$ Telea base fill of $\Omega$; generate two interleaved slit families $\Omega_1$, $\Omega_2$ (width 3 px, 50% duty)

else

  $\Omega_1 \leftarrow \Omega$; $\Omega_2 \leftarrow \emptyset$

for each slit family $\Omega_k \in \{\Omega_1, \Omega_2\}$ do

  $u \leftarrow \nabla\perp g/|\nabla g|$ with $g = G_\sigma \star I$;  $\omega \leftarrow -\Delta I$ inside $\Omega_k$

  for $n = 1, \ldots, N$ do

    $\kappa_\sigma \leftarrow \kappa(|\nabla(G_\sigma \star \omega^n)|)$    (gate, Picard-lagged)

    $\omega^{n+1} \leftarrow \omega^n + dt\cdot[-u\cdot\nabla\omega^n + \nu\Delta\omega^n + \lambda\nabla\cdot(\kappa_\sigma\nabla\omega^n)]$ inside $\Omega_k$

    solve $-\Delta I^{n+1} = \omega^{n+1}$ in $\Omega_k$ with Dirichlet data I on $\Omega_k^c$ (conjugate gradients)

    if 20 divides n then recompute u from the current I

return I

### 3.1 Vorticity–streamfunction formulation

Following Bertalmío, Bertozzi and Sapiro, a gray-scale image I is regarded as the stream function $\psi$ of a two-dimensional incompressible flow, and inpainting a missing region $\Omega$ amounts to advecting the vorticity $\omega$ along the isophote directions to a steady state. With $g = G_\sigma \star I$ a Gaussian-smoothed version of the image, the velocity field is defined as

$$u = \nabla \perp g \,/\, |\nabla g|, \quad \nabla \perp = (-\partial_y, \partial_x) \tag{1}$$

which is unit-length everywhere and tangent to the isophotes. The classical NS inpainting model reads

$$\frac{\partial\omega}{\partial t} + u \cdot \nabla\omega = \nu\Delta\omega, \quad \omega = -\Delta\psi, \quad I = \psi|_\Omega, \tag{2}$$

where $\nu > 0$ is the viscosity. To improve the continuation of high-curvature structures, we add a higher-order correction to the vorticity equation and obtain the proposed model, denoted v0.2:

$$\frac{\partial\omega}{\partial t} + u \cdot \nabla\omega = \nu\Delta\omega + \lambda\nabla \cdot (\,\kappa(|\nabla(G_\sigma \star \omega)|)\,\nabla\omega\,), \tag{3}$$

$$\kappa(s) = (s/K)^2 \,/\, (1 + (s/K)^2), \tag{4}$$

where $\lambda \geq 0$ is the weight of the correction and $K > 0$ a gradient threshold. Since $\omega = -\Delta\psi$, the transport term is of third order and the correction term of fourth order in $\psi$; for $\lambda \to 0$ the model degenerates to (2), which serves as the ablation baseline.

The gate $\kappa$ activates the correction only where the vorticity gradient is significant, that is, near structural discontinuities of the isophote field, and vanishes in flat regions, where the classical model remains. Estimating the argument of $\kappa$ from the smoothed vorticity $G_\sigma \star \omega$, in the spirit of Catté et al., suppresses noise-driven spurious activation; in our experiments this single modification removes the sensitivity of the results to $\lambda$ over the range [0.05, 0.20].

### *3.1.1 The sign dichotomy: a well-posedness analysis*

The correction term is quasi-linear and second order in $\omega$. Its qualitative effect is determined by the signs of its directional diffusion coefficients, which we now compute exactly.

***Definition 1 (directional frame).*** *At points where $\nabla\omega \neq 0$ let $\eta = \nabla\omega/|\nabla\omega|$ be the vorticity-gradient direction and $\xi = \eta\perp$ the orthogonal (iso-vorticity) direction; define the directional second derivatives $\omega_{\eta\eta} := \eta^T D^2\omega\eta$ and $\omega_{\xi\xi} := \Delta\omega - \omega_{\eta\eta}$.*

***Lemma 1 (anisotropic expansion).*** *For $\kappa \in C^1([0,\infty))$, the correction term admits the exact decomposition (5) below, with $\Phi(s) := \kappa(s) + s\cdot\kappa'(s)$. For the gate (4), $\kappa(s) = s^2/(K^2+s^2)$ and $\kappa'(s) = 2sK^2/(K^2+s^2)^2 \geq 0$, hence $\Phi(s) = s^2(3K^2+s^2)/(K^2+s^2)^2 \geq 0$.*

$$\nabla \cdot (\kappa(|\nabla\omega|)\nabla\omega) = \kappa(|\nabla\omega|) \cdot \omega_{\xi\xi} + \Phi(|\nabla\omega|) \cdot \omega_{\eta\eta}, \quad \Phi(s) := \kappa(s) + s\kappa'(s) \tag{5}$$

*Proof.* Expanding the divergence gives

$$\nabla \cdot (\kappa \nabla \omega) = \kappa \Delta \omega + \kappa'(\nabla |\nabla \omega| \cdot \nabla \omega).$$

Wherever $\nabla\omega \neq 0$ one has

$$\nabla |\nabla \omega| = D^2\omega \cdot \eta, \ \nabla |\nabla \omega| \cdot \nabla \omega = |\nabla \omega| \cdot \eta^{\mathrm{T}} D^2 \omega \eta = |\nabla \omega| \cdot \omega_{\eta\eta}.$$

Substituting $\Delta\omega = \omega_{\xi\xi} + \omega_{\eta\eta}$ from Definition 1 then gives (5). The closed form of $\Phi$ follows from the derivative of the gate,

$$\kappa'(s) = \frac{2sK^2}{(K^2+s^2)^2}, \ \Phi(s) = \kappa(s) + s\kappa'(s) = \frac{s^2(3K^2+s^2)}{(K^2+s^2)^2},$$

by direct simplification. ■

Along the two principal directions the right-hand side of (3) therefore acts as a direction-dependent diffusion: the coefficient of $\omega_{\xi\xi}$ is $\nu + \lambda\kappa$ for the forward sign and $\nu - \lambda\kappa$ for the backward sign, and the coefficient of $\omega_{\eta\eta}$ is $\nu + \lambda\Phi$ and $\nu - \lambda\Phi$ accordingly. The signs of these two coefficients decide well-posedness.

***Theorem 1 (ill-posedness of the backward form).*** *Consider the backward variant of (3) (sign $-\lambda$) linearized at a smooth state $\omega_0$ with $|\nabla\omega_0| = s_0$. If $\kappa(s_0) > \nu/\lambda$, equivalently if $(\nu,\lambda,K,s_0)$ satisfy $s_0^2 > K^2(\nu/\lambda)/(1-\nu/\lambda)$, then the frozen-coefficient principal operator has a negative diffusion coefficient and the Cauchy problem is ill-posed in the sense of Hadamard: for every $t > 0$ the amplification factor of plane waves is $exp((\lambda\kappa(s_0)-\nu)|k|^2 t)$, which is unbounded as $|k| \to \infty$, so the solution (whenever it exists) does not depend continuously on the initial data in any Sobolev norm. With $(\nu, \lambda, K) = (0.02, 0.05, 0.02)$ the threshold reads $s_0 \gtrsim 0.016$, which is attained at structural edges of real images after the Telea initialization, where measured $|\nabla\omega|$ reaches $O(0.1)$.*

*Proof.* Freezing $\kappa$ at $s_0$, the $\xi$-directional part of the linearized equation reads

$$\frac{\partial w}{\partial t} = (\nu - \lambda\kappa(s_0)) \cdot w_{\xi\xi}, \ \nu - \lambda\kappa(s_0) < 0,$$

into which the plane waves

$$w_k(x,t) = exp((\lambda\kappa(s_0) - \nu)|k|^2 t) \cdot e^{ik\cdot x}, \ k \parallel \xi,$$

substitute as exact solutions, with amplification factor

$$\|w_k(t)\|_2 / \|w_k(0)\|_2 = exp((\lambda\kappa(s_0) - \nu)|k|^2 t) \to \infty \ \ as\, |k| \to \infty,$$

for every $t > 0$, violating Hadamard's continuous-dependence requirement in every Sobolev space $H^s$. The same computation with $\Phi$ in place of $\kappa$ covers the $\eta$ direction. ■

Two clarifications are in order. Smoothing the gate argument to $\kappa_\sigma = \kappa(|\nabla(G_\sigma \star \omega)|)$ does

not cure the backward form: it regularizes the coefficient $\kappa_\sigma$, but the operator $-\lambda\kappa_\sigma\Delta\omega$ retains the positive frozen symbol $+(\lambda\kappa_\sigma - \nu)|k|^2$. The sign, not the gate estimate, is the source of ill-posedness, in agreement with the experiments, where the backward form loses 0.02 dB although it uses the same gate. Nor can the forward form enter the backward regime: $\kappa \geq 0$ and $\Phi \geq 0$ by Lemma 1, so both directional coefficients satisfy $\nu + \lambda\kappa \geq \nu > 0$ and $\nu + \lambda\Phi \geq \nu > 0$, and the operator is uniformly parabolic.

***Lemma 2 (smoothing bound).*** *For any $\omega \in L^2(\Omega)$, $\|\nabla(G_\sigma \star \omega)\|_{L^\infty} \leq C_\sigma\|\omega\|_{L^2}$ with $C_\sigma := \|\nabla G_\sigma\|_{L^2} < \infty$; hence the smoothed gate $\kappa_\sigma(x) := \kappa(|\nabla(G_\sigma \star \omega)|)$ takes values in [0,1] and depends on $\omega$ only through a lower-order ($L^2$) quantity.*

*Proof.* Apply Young's convolution inequality to

$$\nabla(G_\sigma \star \omega) = (\nabla G_\sigma) \star \omega, \quad \|\nabla(G_\sigma \star \omega)\|_{L^\infty} \leq \|\nabla G_\sigma\|_{L^2} \cdot \|\omega\|_{L^2},$$

which is exactly the claimed bound with $C_\sigma := \|\nabla G_\sigma\|_{L^2} < \infty$. ■

***Theorem 2 (energy stability of the forward model).*** *Let $\omega$ solve the forward model (3) with frozen smoothed gate $\kappa_\sigma$ (Picard step), homogeneous Dirichlet data $\omega = 0$ on $\partial\Omega$, and a Lipschitz direction field $u$ with $\|\nabla u\|_{L^\infty} \leq C_u$ (guaranteed by the Gaussian smoothing in (1)). Then for every $t \geq 0$ the energy inequality (6) holds, and consequently $\|\omega(t)\|^2_{L^2} \leq exp(C_u \cdot t) \cdot \|\omega(0)\|^2_{L^2}$.*

$$\frac{d}{dt}\|\omega\|^2 + 2\nu\|\nabla\omega\|^2 + 2\lambda\int_\Omega \kappa_\sigma|\nabla\omega|^2\,dx \leq \|\nabla u\|_\infty \cdot \|\omega\|^2 \qquad (6)$$

*Proof.* Multiply (3) by ω and integrate over Ω. The transport term integrates by parts to

$$\int_\Omega \omega \cdot (u \cdot \nabla\omega)\,dx = -\tfrac{1}{2}\int_\Omega \omega^2(\nabla \cdot u)\,dx \leq \tfrac{1}{2}C_u\|\omega\|^2,$$

where the boundary term vanishes since $\omega = 0$ on $\partial\Omega$. The viscous term gives

$$\int_\Omega \omega \cdot \nu\Delta\omega\,dx = -\nu\|\nabla\omega\|^2$$

by Green's identity, and the correction term gives

$$\int_\Omega \omega \cdot \lambda\nabla \cdot (\kappa_\sigma\nabla\omega)\,dx = -\lambda\int_\Omega \kappa_\sigma|\nabla\omega|^2\,dx \leq 0.$$

Collecting the three estimates yields (6), and Grönwall's lemma gives the claimed bound. ■

***Corollary 1 (consequences).*** *(i) The forward model is $L^2$-stable uniformly in $\lambda \geq 0$; applying (6) to the difference of two solutions of the frozen-gate model yields uniqueness and continuous dependence on the data. (ii) For the backward sign the correction enters the energy identity with the opposite sign, $+\lambda\int \kappa_\sigma|\nabla\omega|^2$, which can be absorbed neither by $\nu\|\nabla\omega\|^2$ (since $\kappa_\sigma \leq 1$* *does not control the ratio) nor by $\|\omega\|^2$ (the Rayleigh quotient $\int \kappa_\sigma|\nabla\omega|^2/\|\omega\|^2$ is unbounded as $\omega$ concentrates near max $\kappa_\sigma$): no a priori energy estimate exists, in agreement with Theorem 1. (iii) Local-in-time well-posedness of the quasi-linear forward model with smoothed gate follows from standard uniformly parabolic theory; the argument of Catté et al. for the Perona–Malik equation transfers verbatim, since by Lemma 2 the coefficient $\kappa_\sigma$ depends only on a lower-order norm of $\omega$. Long-time regularity and the unsmoothed quasi-linear case remain open problems.*

Controlled experiments corroborate the dichotomy: on identical crack-mask configurations the backward form loses 0.02 dB consistently, whereas the forward form gains monotonically in $\lambda$ and, with the smoothed gate, stays flat over $\lambda \in [0.05, 0.20]$.

### *3.1.2 The order–dissipation principle*

Fourier symbols determine which orders of correction are admissible at all. This explains why the model carries a fourth-order correction and why a fifth-order term in $\psi$, for instance, cannot serve as a regularizer.

***Lemma 3 (parity of symbols).*** *Let $P = \Sigma_{|\alpha|\leq m} c_\alpha \partial^\alpha$ be a constant-coefficient operator with real coefficients and principal part $P_m$. Its symbol $\sigma(k) = \Sigma\, c_\alpha (ik)^\alpha$ satisfies: (i) if m is odd, $Re\,\sigma_m(k) \equiv 0$; (ii) if m = 2j is even, $\sigma_m(k)$ is real and homogeneous of degree 2j. The semigroup $e^{tP}$ is $L^2$-dissipative, i.e., $\|e^{tP}w\|_{L^2} \leq \|w\|_{L^2}$ for all data, only if $Re\,\sigma(k) \leq 0$ for all k; this forces m even and $\sigma_m(k) \leq 0$, i.e., a principal part of $(-1)^{j+1}(-\Delta)^j$ type. For odd m neither sign yields dissipation, since $Re(\pm\sigma_m) \equiv 0$.*

*Proof.* Under the Fourier transform, the equation and its energy rate read

$$\frac{\partial \hat{w}}{\partial t} = \sigma(k)\hat{w}, \quad \frac{d}{dt}|\hat{w}|^2 = 2Re\,\sigma(k)\cdot|\hat{w}|^2.$$

With real coefficients $c_\alpha$, $(ik)^\alpha$ contributes to Re $\sigma$ exactly when $|\alpha|$ is even; an odd principal part therefore has a purely imaginary symbol. Dissipation for all data is equivalent to

$$Re\,\sigma(k) \leq 0 \quad for\ all\ k,$$

which is impossible for a nonzero odd-order homogeneous symbol, an odd function of k, and for an even-order symbol requires the stated sign. ■

***Proposition 1 (design consequences).*** *In the $\psi$-caliber ($\omega = -\Delta\psi$): (i) the transport $u\cdot\nabla\omega = -u\cdot\nabla\Delta\psi$ is third order, hence skew and dispersive: it redistributes vorticity along isophotes without dissipating energy and has a Hamiltonian character; (ii) the viscosity $\nu\Delta\omega = -\nu\Delta^2\psi$ is fourth order with symbol $-\nu|k|^4$, dissipative; (iii) the correction $\lambda\nabla\cdot(\kappa\nabla\omega)$ is fourth order and* *dissipative exactly in its forward form (Theorem 1); (iv) any fifth-order correction in $\psi$ has a purely imaginary principal symbol and cannot regularize, regardless of sign — such terms are inadmissible as regularizers; (v) the next admissible even order is the sixth, realized by the triharmonic term $-\mu\Delta^2\omega$ with symbol $-\mu|k|^6$, which we use as the upper-limit reference in the order ablation; intermediate orders are available only through fractional Laplacians $(-\Delta)^s$, $s \in (1, 3/2)$.*

Proposition 1 turns the order selection from an empirical choice into a design principle: the proposed model is the lowest-order system that combines a third-order transport mechanism with a fourth-order dissipative anisotropic regularization, and the gate κ is what distinguishes it from plain biharmonic diffusion, the linear fourth-order alternative.

### 3.2 Numerical scheme

The model is solved by an explicit vorticity marching coupled with a Poisson reconstruction of the image. Let H denote the characteristic function of the hole Ω and let $I^0$ be an initial guess obtained by the fast marching method of Telea. At each step n:

**(a)** Transport: $-u\cdot\nabla\omega$ is discretized by a first-order upwind scheme, which preserves monotonicity; the direction field u is recomputed from the current image every 20 steps (the u–ψ coupling).

**(b)** Viscous and correction terms: $\nu\Delta\omega$ is discretized by the five-point Laplacian; the correction term is Picard-lagged, with κ evaluated at $\omega^n$ and the divergence discretized by central differences.

**(c)** Vorticity update: $\omega^{n+1} = \omega^n + dt\cdot(\text{rhs})\cdot H$; only unknowns inside Ω are advanced.

**(d)** Poisson reconstruction: $I^{n+1}$ solves $\Delta I^{n+1} = \Delta I^n + (\omega^{n+1} - \Delta I^n_{lum})\cdot H$ in Ω with Dirichlet data $I^n$ on $\bar{\Omega}^c$. Since the five-point Laplacian is negative semi-definite, the conjugate-gradient iteration is applied to the SPD form $-\Delta\psi = -\text{rhs}$, and the matrix–vector product involves only the linear part $A_0(p) = -\Delta(pH)H$; mixing the affine boundary contribution into the iteration matrix destroys convergence.

With dt = 0.02 and ν = 0.02 the scheme is stable, and accuracy peaks at about 50 iterations, with PSNR 36.9, 37.5 and 36.8 at 20, 50 and 100 steps on a validation configuration, which reflects the balance between transport progress and accumulated discretization error.

### 3.3 Stripe decomposition for large missing blocks

Inside a large missing block the vorticity field is essentially flat after the initial filling: the measured activation rate of κ is only 7.3% for block masks, against 56% for crack masks. The higher-order term is therefore silent on block damage, and all configurations collapse onto the Telea baseline. Stripe decomposition converts the block problem into a sequence of slit problems, in which the correction term becomes active again. The procedure has three steps.

**(i)** Base filling: the whole block is first inpainted by Telea, providing a plausible known background. This step is essential because inpainting solvers treat every pixel outside the current mask as known data, and unfilled black gaps would otherwise be pulled into the slits as false observations.

**(ii)** Stripe masks: parallel stripes of width w = 3 px and duty cycle 50% are generated; the stripe normal is aligned with the dominant gradient direction estimated from the structure tensor of the known ring around the block, or fixed horizontally, which performs equivalently.

**(iii)** Interleaved passes: the stripes are inpainted in two phase-shifted passes. In each pass only one family of slits is unknown, both sides of every slit already contain data (original or previously filled), and the higher-order NS solver operates exactly in its favorable regime. The union of the passes covers the whole block.

The strategy is base-agnostic: applied to Telea or to the classical NS solver it yields a consistent gain of +0.16 to +0.22 dB on 40 controlled BSDS500 block samples at 20% and 30% area, and combined with the forward higher-order term the gains add up to +0.45 dB over the direct classical NS solver.

### 3.4 Implementation and parameterization

The solver is implemented in PyTorch and runs on a consumer GPU, an RTX 3060 Laptop with 6 GB of memory. A 512 × 512 image with a 15% crack mask takes 4–16 s depending on the configuration. Table 1 lists the default parameters; the model is insensitive to λ and requires no per-image tuning.

**Table 1 Default parameters of model v0.2**

| Parameter | Value | Role |
|---|---|---|
| dt, ν | 0.02, 0.02 | time step; viscosity |
| iters | 50 | vorticity marching steps (accuracy optimum) |
| λ, K | 0.05, 0.02 | correction weight; gate threshold (insensitive over one order) |
| σ (κ smoothing) | 1.0 | Catté-type smoothing inside the gate κ |
| σ (velocity) | 1.5 | smoothing for the isophote direction field |
| w, d | 3 px, 3 px | stripe width and spacing (two interleaved passes) |

## 4. Experiments

## 4.1 Setup

Datasets. Three sources are used: the BSDS500 test set of natural images at 481 × 321 resolution; the DIV2K validation set, 100 images rescaled to 512 px on the long side; and a classical Chinese painting restoration benchmark with a test split of 42 images at 512 × 512, whose fine ink strokes provide a stringent structural testbed.

Damage protocols. Three mask families with fixed seeds are used throughout: structural cracks consisting of 4 random polylines traversing the image with width 8–12 px and a mask ratio of about 14.6%, random blocks covering 20% or 30% of the area, and programmatic text occlusion. Metrics are PSNR, SSIM and LPIPS.

Baselines. Twelve methods are compared. The classical group comprises Telea fast marching, the OpenCV implementation of the classical NS model, and our reproductions of CDD and the Criminisi exemplar-based method; the latter costs 13 min per 512 × 512 image in our faithful full-search implementation and is evaluated on a 6-image subset per scenario. The recent PDE group, published between 2023 and 2026 and reproduced by us under a unified interface, comprises the double-CDD model with p-Laplace operator, the Ginzburg–Landau model with SAV algorithm, fractional-Laplacian regularization, curvature-dependent elastic-bending TV, and the accelerated operator-splitting solver of Euler's elastica. LaMa with the official checkpoint serves as the learning-based reference, inference only. The diffusion-based RePaint is evaluated separately on a 256 × 256 subset of 30 samples in Section 4.10, because its 550 s per image on our GPU exceeds the budget of the full benchmark. All PDE solvers run on an RTX 3060 Laptop GPU with 6 GB of memory; the proposed solver costs 4–16 s per image.

## 4.2 Structural cracks: the solver advantage

On crack masks the proposed vorticity–streamfunction solver, already with $\lambda = 0$, outperforms both classical transport baselines by a wide margin, Table 2, rows 1–4. On the full 42-image painting set, Table 3, model v0.2 beats all five classical baselines on every sample, gaining +1.02 dB over Telea with a 42/42 win rate and 40/42 against its own $\lambda = 0$ variant; the sorted per-sample gains in Figure 2 show that the advantage is uniform rather than outlier-driven. Against the strongest recent PDE solvers the margin narrows: the fractional-Laplacian model and the Euler-elastica operator splitting lead PSNR by 0.36 and 0.31 dB respectively, while v0.2 attains the best LPIPS among all eleven PDE-based methods, 0.062 against 0.073 for the two leaders, and surpasses the learning-based LaMa in PSNR by 0.18 dB with 28 wins in 42 samples; LaMa leads LPIPS at 0.024. Figure 1 shows a representative painting sample: the transport-type solvers continue the ink strokes across the cracks, whereas CDD blurs them, Criminisi copies mismatched texture, and GL-SAV over-smooths; the magnified insets of the marked foliage region make these differences directly visible. Figure 3 summarizes all twelve methods.

**Table 2 Crack damage, BSDS and painting subsets (mask ratio ≈ 14.6%)**

| Dataset / Method | PSNR↑ | SSIM↑ | LPIPS↓ |
|---|---|---|---|
| BSDS / Telea | 28.259 | 0.9147 | 0.0725 |
| BSDS / NS-OpenCV | 28.172 | 0.9157 | 0.0780 |
| BSDS / ours (λ=0) | **29.380** | **0.9314** | **0.0588** |
| BSDS / backward λ=0.05 | <u>29.355</u> | <u>0.9311</u> | <u>0.0589</u> |
| Painting / Telea | 30.178 | 0.9290 | 0.0711 |
| Painting / NS-OpenCV | 30.225 | 0.9314 | 0.0740 |
| Painting / ours (λ=0) | **30.961** | **0.9403** | **0.0612** |
| Painting / backward λ=0.05 | <u>30.945</u> | <u>0.9402</u> | <u>0.0612</u> |

**Table 3 Crack damage, full 42-image painting test set (sorted by PSNR)**

| Method | PSNR↑ | SSIM↑ | LPIPS↓ |
|---|---|---|---|
| Fractional Laplacian | **32.977** | **0.9439** | 0.0733 |
| Euler elastica OS | <u>32.926</u> | <u>0.9435</u> | 0.0731 |
| v0.2 (ours) | 32.614 | 0.9426 | <u>0.0618</u> |
| LaMa | 32.433 | 0.9428 | **0.0243** |
| NS direct (λ=0) | 32.387 | 0.9414 | 0.0666 |
| Telea | 31.596 | 0.9312 | 0.0760 |
| NS-OpenCV | 31.552 | 0.9318 | 0.0814 |
| Elastica-TV | 31.216 | 0.8934 | 0.0992 |
| Double-CDD p-Laplace | 28.665 | 0.8202 | 0.1988 |
| CDD | 26.811 | 0.8392 | 0.1892 |
| GL-SAV | 22.742 | 0.5638 | 0.4163 |
| Criminisi (6-img subset) | 24.710 | 0.8671 | 0.0911 |

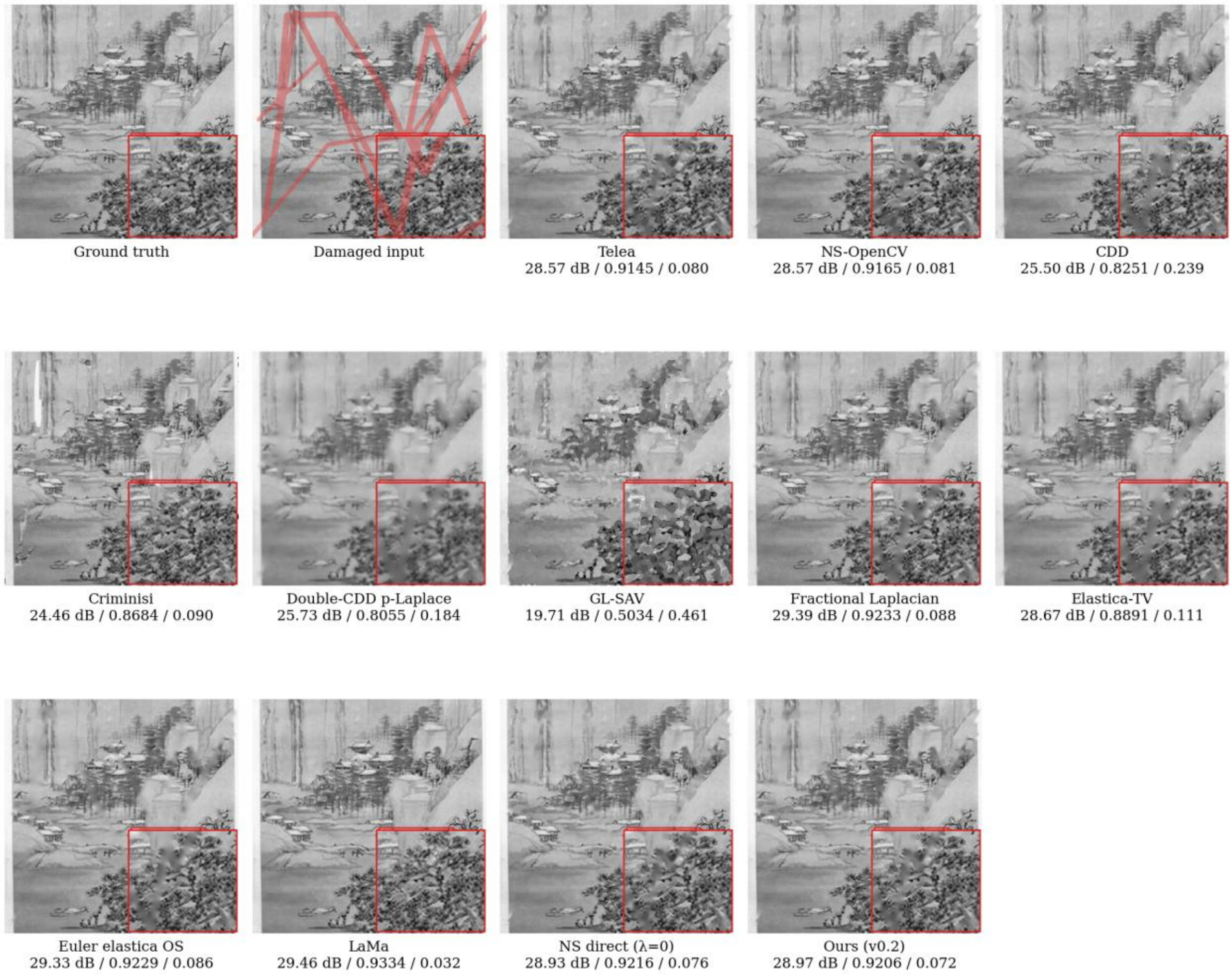

**Figure 1 Crack inpainting on a classical painting (mask ratio 16.9%). Numbers beneath each method: PSNR (dB) / SSIM / LPIPS on this sample. The red rectangle marks the region magnified in the corner inset of every panel, where cracks cross the dense ink foliage.**

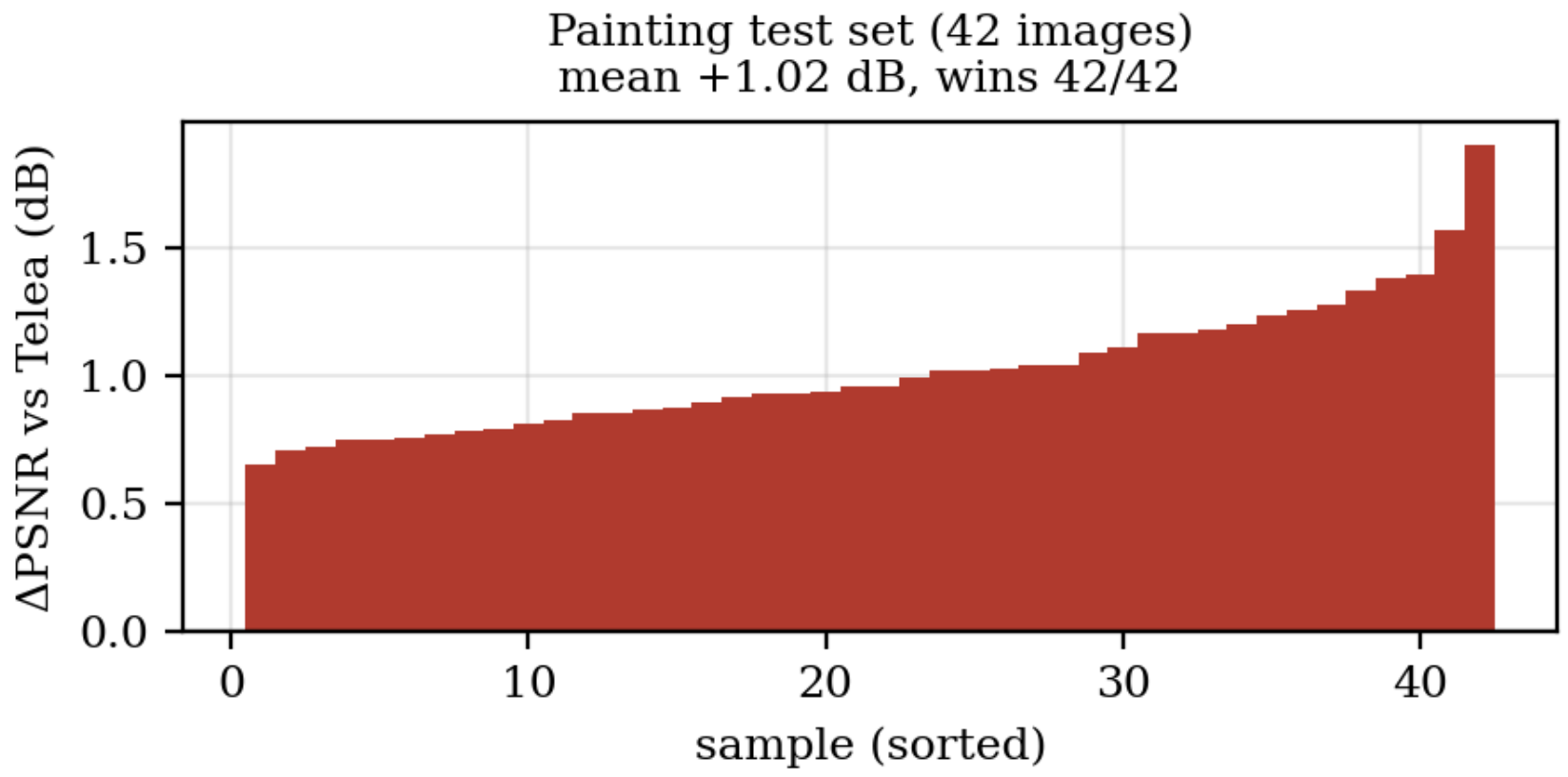

**Figure 2 Per-sample PSNR gain of v0.2 over Telea on the full 42-image painting test set, sorted. Every sample improves; mean +1.02 dB.**

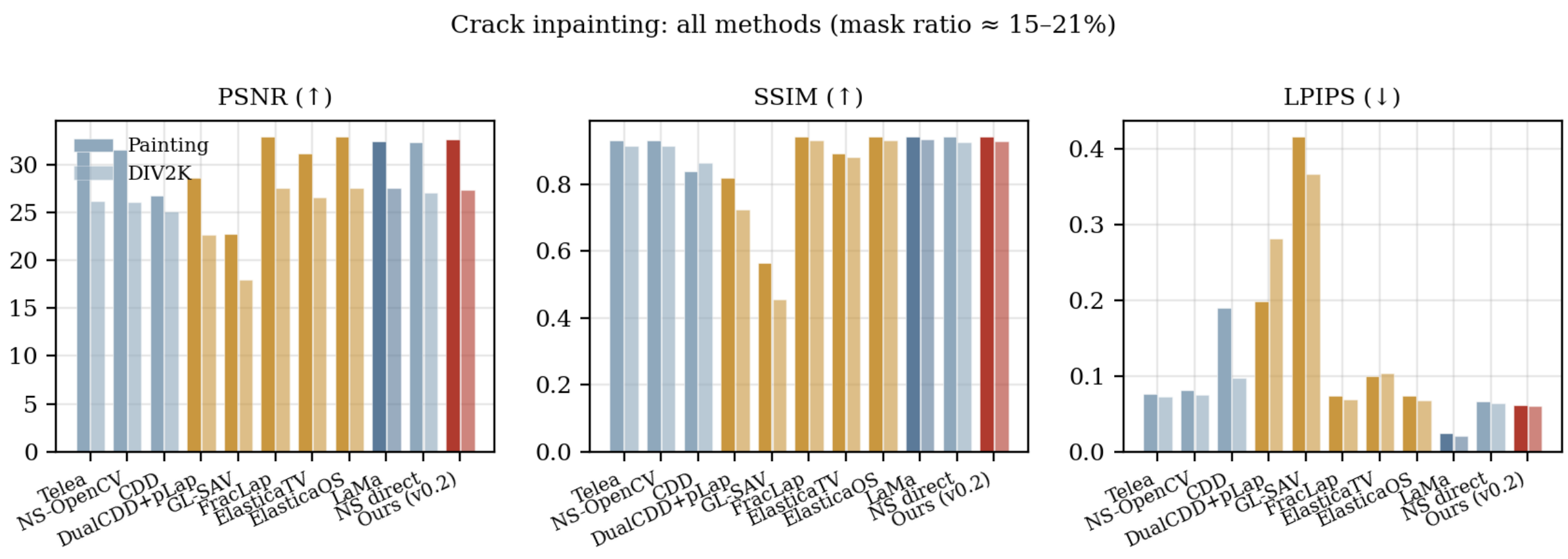

**Figure 3 Crack inpainting, all twelve methods (42 painting + 100 DIV2K images): mean PSNR and LPIPS. v0.2 beats every classical baseline on both datasets and attains the best LPIPS among PDE-based methods.**

### 4.3 Sign ablation: experimental evidence for Theorem 1

Nine crack samples, five from BSDS and four from the paintings, were processed with the correction term in its backward and forward forms, Table 4; the backward form is the one the original draft of this work used. The backward form degrades the result at every weight tested, as the ill-posedness analysis predicts, and the forward form improves monotonically in $\lambda$. The ablation isolates the sign rather than the gate as the deciding factor, which corroborates Theorem 1 and the observation that smoothing alone cannot save the backward form.

**Table 4 Sign-direction ablation (mean PSNR, crack masks)**

| Configuration | BSDS | Painting |
|---|---|---|
| $\lambda = 0$ (no correction) | 29.502 | 29.172 |
| backward $\lambda = 0.05$ | 29.480 | 29.159 |
| forward $\lambda = 0.02$ | 29.509 | 29.175 |
| forward $\lambda = 0.05$ | 29.520 | 29.181 |
| forward $\lambda = 0.10$ | **29.535** | **29.188** |

A companion ablation varies the order itself on 12 crack samples, six from BSDS and six from the paintings with fixed seeds, and Figure 4 reports paired per-sample PSNR differences against the second-order model. The backward fourth-order correction loses 0.02 dB on BSDS and 0.01 dB on the paintings, while the forward correction gains up to +0.03 dB on BSDS and is neutral-to-positive on the paintings, the direction Theorem 1 predicts. A sixth-order term, alone or combined with the fourth, adds at most +0.01 dB over the forward fourth order, far below the gain of getting the sign right; the fourth order is therefore both the lowest admissible one, by Proposition 1, and the practically sufficient one. The ordering of all configurations is identical on both datasets, so the conclusions do not depend on the image source.

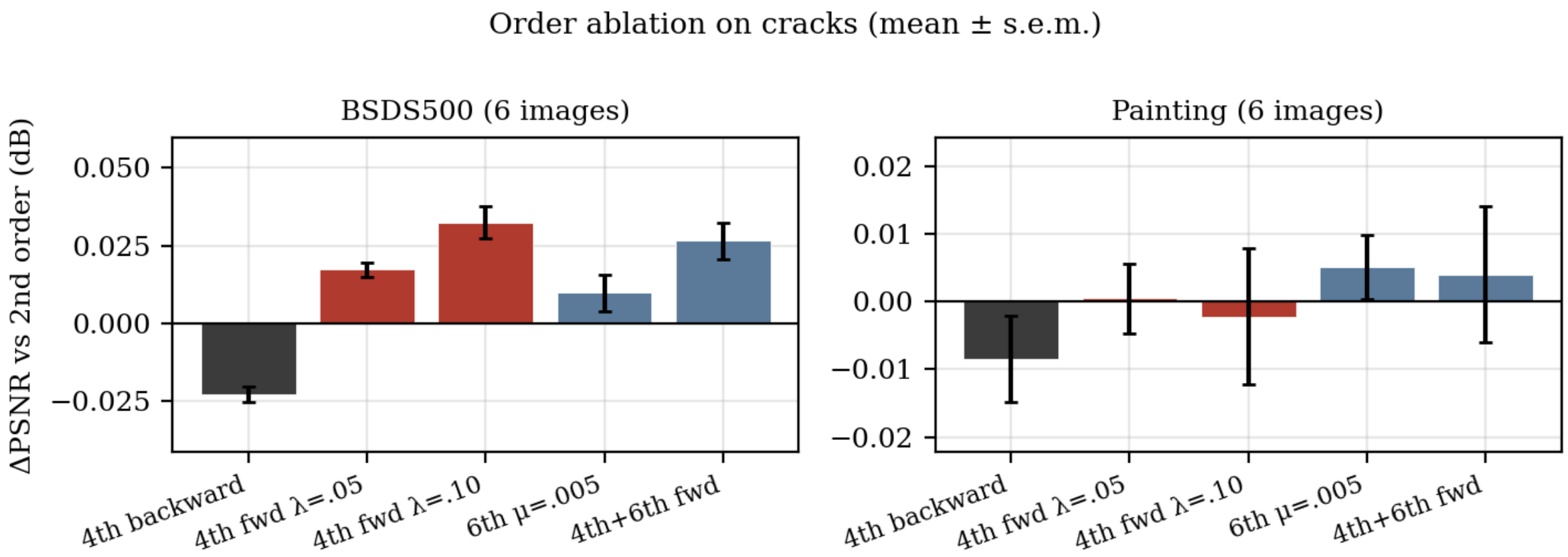

**Figure 4 Order ablation on cracks (12 samples, fixed seeds): paired per-sample PSNR differences against the second-order model (zero line), mean ± s.e.m. The backward fourth order (black) degrades both datasets; the forward fourth order (red) improves them; a sixth-order term adds nothing beyond the fourth.**

### 4.4 Block damage: gate silence and stripe decomposition

On 20% and 30% block masks all direct NS configurations collapse onto the Telea baseline, with differences below 0.01 dB. The cause is visible in the gate itself: after the Telea initialization the intra-hole vorticity gradient is tiny, with mean $|\nabla\omega|$ around 0.009, so with $K = 0.02$ the gate $\kappa$ activates on only 7.3% of hole pixels, against 56% on cracks, and the higher-order term stays silent where there is no structure to continue. Stripe decomposition reactivates it by converting the block into a sequence of slits. Table 5 and Figure 6 show the effect to be base-agnostic, +0.16 dB when wrapping Telea and +0.22 dB when wrapping the classical NS solver, with the stripe version better on every one of the 40 samples. Figure 5 shows a representative block example: the stripe-based fill restores the vertical contour of the figure and the water line, visible in the magnified insets of the marked region, while the direct solvers leave a visible blur and the exemplar-based method misplaces structure.

**Table 5 Block damage (BSDS, 40 samples): stripe decomposition vs direct solvers**

| Method | PSNR↑ | SSIM↑ | LPIPS↓ |
|---|---|---|---|
| stripe-Telea (horizontal) | **21.567** | 0.8257 | **0.1540** |
| stripe-NS (λ=0, horizontal) | <u>21.514</u> | **0.8266** | 0.1608 |
| stripe-NS (λ=0, adaptive dir.) | 21.499 | <u>0.8265</u> | <u>0.1592</u> |
| Telea | 21.406 | 0.8255 | 0.1659 |
| NS-OpenCV | 21.359 | 0.8240 | 0.1696 |
| ours direct (λ=0) | 21.296 | 0.8256 | 0.1696 |

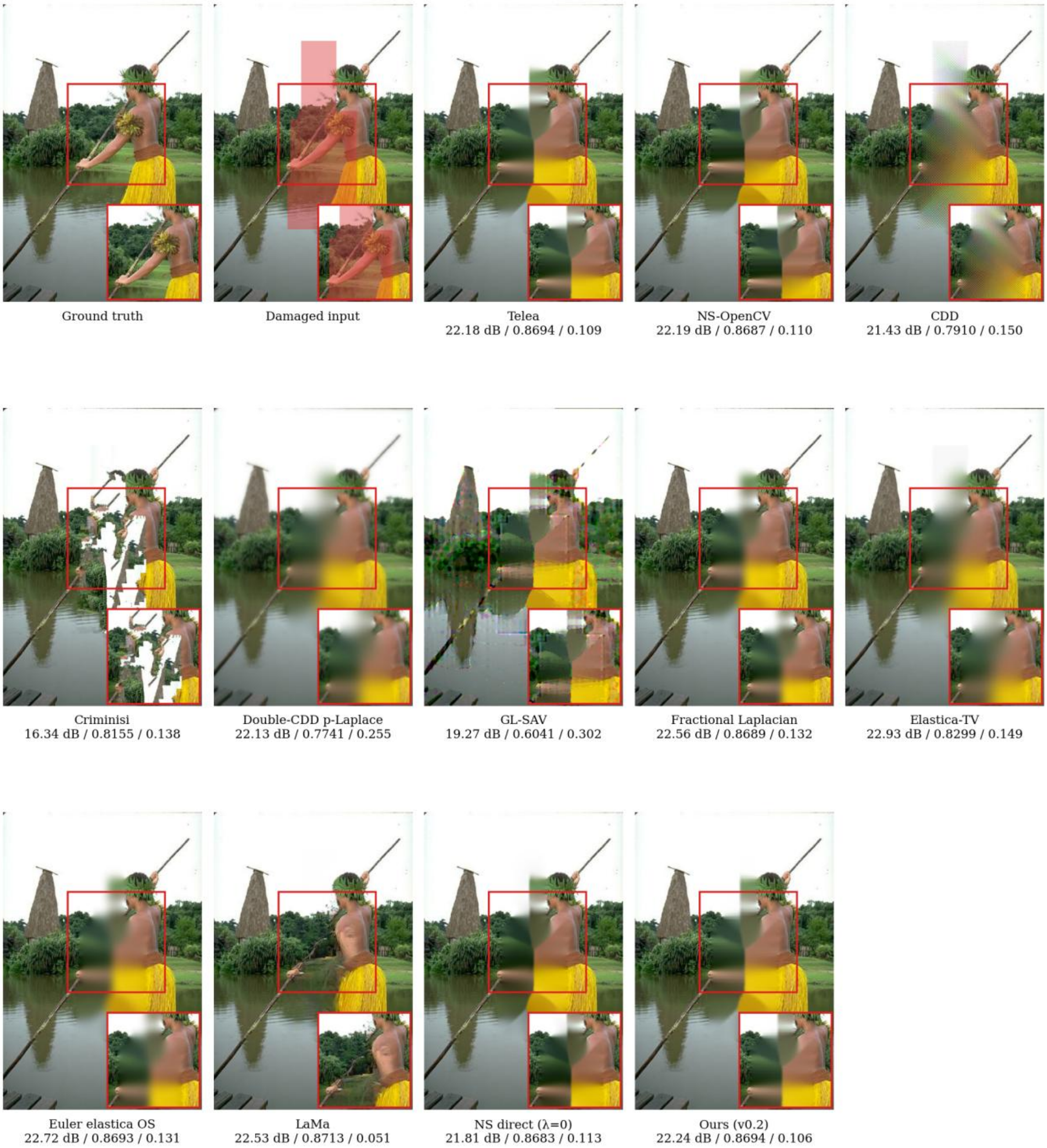

**Figure 5 Block inpainting (20% area, BSDS500). Numbers beneath each method: PSNR (dB) / SSIM / LPIPS on this sample. The red rectangle marks the region magnified in the corner inset of every panel, covering the vertical contour of the figure and the water line.**

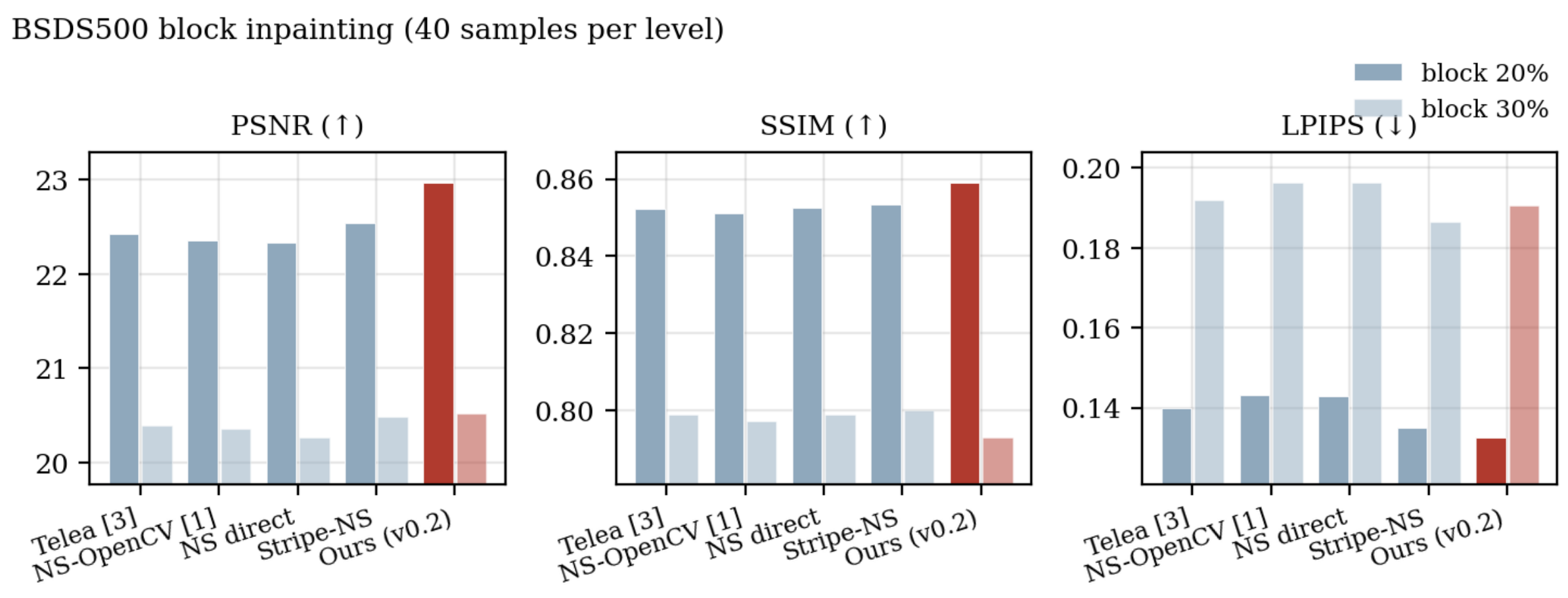

**Figure 6  Block inpainting on BSDS500 (40 samples per level): PSNR, SSIM and LPIPS of the direct baselines, stripe-NS and the combined model v0.2 at 20% and 30% damage.**

### 4.5 The combined model: additive gains and parameter insensitivity

Combining stripe decomposition with the forward correction, model v0.2, stacks the two gain sources, Table 6: +0.45 dB over the direct classical solver and +0.34 dB over Telea on BSDS blocks. With the smoothed gate the PSNR varies by at most 0.002 dB over $\lambda \in [0.05, 0.20]$, so no per-image tuning is needed. The average hides considerable heterogeneity, with per-sample differences against stripe-Telea ranging from −3.1 to +4.2 dB, mean +0.18 dB and a 22/40 win rate. We also tested whether five cheap structural features, namely ring gradient strength, edge density, tensor coherence, gate-activation rate and mask ratio, can predict the sign of the per-sample gain. They cannot: correlations stay below $|r| = 0.25$, and leave-one-out gating gains at most +0.08 dB, at the noise edge. We therefore recommend the uniform forward configuration rather than a learned or heuristic gate.

**Table 6  Combined model on BSDS blocks (40 samples)**

| Method | PSNR↑ | SSIM↑ | LPIPS↓ |
|---|---|---|---|
| v0.2, λ=0.05 | **21.744** | <u>0.8259</u> | 0.1615 |
| v0.2, λ=0.10 | <u>21.743</u> | 0.8259 | 0.1616 |
| v0.2, λ=0.20 | 21.742 | 0.8259 | 0.1617 |
| v0.2, adaptive dir. | 21.731 | 0.8259 | <u>0.1607</u> |
| stripe-Telea | 21.567 | 0.8257 | **0.1540** |
| stripe-NS (λ=0) | 21.514 | **0.8266** | 0.1608 |
| Telea | 21.406 | 0.8255 | 0.1659 |

### 4.6 Full painting-set evaluation and additional baselines

On painting blocks, Table 7, the ordering changes character: the learning-based LaMa leads all three metrics by a wide margin, at 27.649 dB PSNR and 0.055 LPIPS, and among PDE solvers the recent fourth-order models edge ahead, while v0.2 ties NS-OpenCV in PSNR, 26.744 against 26.745, and keeps the best LPIPS among the PDE family at 0.130. The stripe gain is narrower than on BSDS, +0.06 dB over Telea with 37 wins in 42 samples, because the fine painting texture leaves less room for structural refinement. Two baseline observations are worth reporting. Our CDD reproduction is far weaker than the transport-type solvers on both scenarios, which is consistent with the view that CDD suits thin scratches rather than wide cracks or blocks and supports the value of a transport-type higher-order route. The Criminisi reproduction ranks last in PSNR yet second-best in LPIPS among PDE methods at 0.112: exemplar copying produces perceptually sharp but spatially inaccurate content, so multi-baseline comparisons should be reported per metric rather than by PSNR alone. Figure 7 collects the block results across both datasets.

**Table 7 Block damage 20%, full 42-image painting test set (sorted by PSNR)**

| Method | PSNR↑ | SSIM↑ | LPIPS↓ |
|---|---|---|---|
| LaMa | **27.649** | 0.8821 | **0.0554** |
| Euler elastica OS | <u>27.051</u> | **0.8942** | 0.1499 |
| Fractional Laplacian | 26.935 | <u>0.8939</u> | 0.1519 |
| Elastica-TV | 26.765 | 0.8372 | 0.1575 |
| NS-OpenCV | 26.745 | 0.8903 | 0.1416 |
| v0.2 (ours) | 26.744 | 0.8924 | 0.1296 |
| Telea | 26.683 | 0.8915 | 0.1320 |
| NS direct (λ=0) | 26.477 | 0.8921 | 0.1376 |
| Double-CDD p-Laplace | 25.737 | 0.7891 | 0.2682 |
| CDD | 23.271 | 0.7808 | 0.1929 |
| GL-SAV | 21.559 | 0.5786 | 0.4110 |
| Criminisi (6-img subset) | 20.581 | 0.7869 | <u>0.1115</u> |

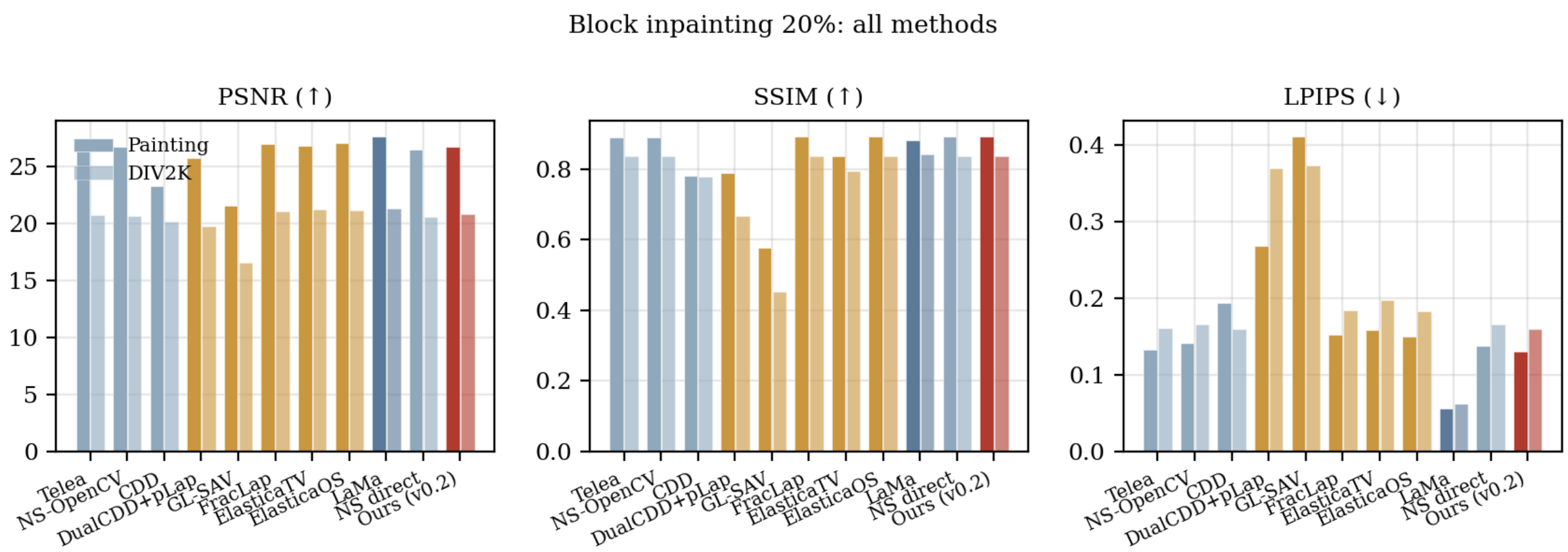

**Figure 7  Block inpainting (20% area), all twelve methods (42 painting + 100 DIV2K images): mean PSNR and LPIPS. LaMa leads on large semantic holes; v0.2 ties the classical transport solvers in PSNR and keeps the best LPIPS among PDE-based methods.**

Positioning. The evidence supports a precise claim: on structure-dominated thin-to-medium damage such as cracks and scratches, the proposed model is the strongest PDE solver in perceptual quality, beating every classical baseline on every sample and matching the strongest recent PDE solvers within 0.4 dB PSNR while leading them all on LPIPS; on large blocks it offers PSNR on par with the best classical baseline and the best perceptual quality among PDE methods, while semantic completion of large holes remains the territory of learning-based methods such as LaMa. Structure-dominated damage is the regime of the motivating application, the restoration of cracks and scratches in classical paintings.

### 4.7 DIV2K validation set

The full DIV2K validation set, 100 images rescaled to 512 px on the long side and 200 samples across the two scenarios, confirms the conclusions on an independent source of natural images; Table 8 covers cracks and Table 9 blocks. On cracks v0.2 beats Telea and NS-OpenCV on all 100 samples, by +1.17 and +1.22 dB respectively as Figure 8 shows, and again attains the best LPIPS among PDE-based methods at 0.061; the fractional-Laplacian and Euler-elastica solvers lead PSNR by about 0.25 dB, and LaMa leads by 0.27 dB while dominating LPIPS at 0.021. On blocks v0.2 beats Telea on 98 of 100 samples with a mean gain of +0.08 dB and keeps the best LPIPS among PDE methods, while LaMa leads outright at 21.301 dB PSNR and 0.062 LPIPS, the same pattern as on the paintings.

**Table 8 Crack damage, DIV2K validation set (100 samples, sorted by PSNR)**

| Method | PSNR↑ | SSIM↑ | LPIPS↓ |
|---|---|---|---|
| Fractional Laplacian | **27.594** | 0.9309 | 0.0682 |
| LaMa | 27.588 | **0.9341** | **0.0209** |
| Euler elastica OS | 27.560 | 0.9306 | 0.0679 |
| v0.2 (ours) | 27.320 | 0.9289 | 0.0605 |
| NS direct (λ=0) | 27.041 | 0.9271 | 0.0634 |
| Elastica-TV | 26.533 | 0.8819 | 0.1034 |
| Telea | 26.153 | 0.9135 | 0.0728 |
| NS-OpenCV | 26.099 | 0.9143 | 0.0752 |
| CDD | 25.108 | 0.8634 | 0.0966 |
| Double-CDD p-Laplace | 22.704 | 0.7250 | 0.2814 |
| GL-SAV | 17.931 | 0.4552 | 0.3663 |

**Table 9 Block damage 20%, DIV2K validation set (100 samples, sorted by PSNR)**

| Method | PSNR↑ | SSIM↑ | LPIPS↓ |
|---|---|---|---|
| LaMa | **21.301** | **0.8423** | **0.0622** |
| Elastica-TV | 21.213 | 0.7951 | 0.1975 |
| Euler elastica OS | 21.126 | 0.8386 | 0.1823 |
| Fractional Laplacian | 21.024 | 0.8383 | 0.1839 |
| v0.2 (ours) | 20.797 | 0.8373 | 0.1597 |
| Telea | 20.717 | 0.8373 | 0.1604 |
| NS-OpenCV | 20.674 | 0.8362 | 0.1658 |
| NS direct (λ=0) | 20.556 | 0.8366 | 0.1654 |
| CDD | 20.156 | 0.7778 | 0.1589 |
| Double-CDD p-Laplace | 19.754 | 0.6674 | 0.3695 |
| GL-SAV | 16.568 | 0.4521 | 0.3723 |

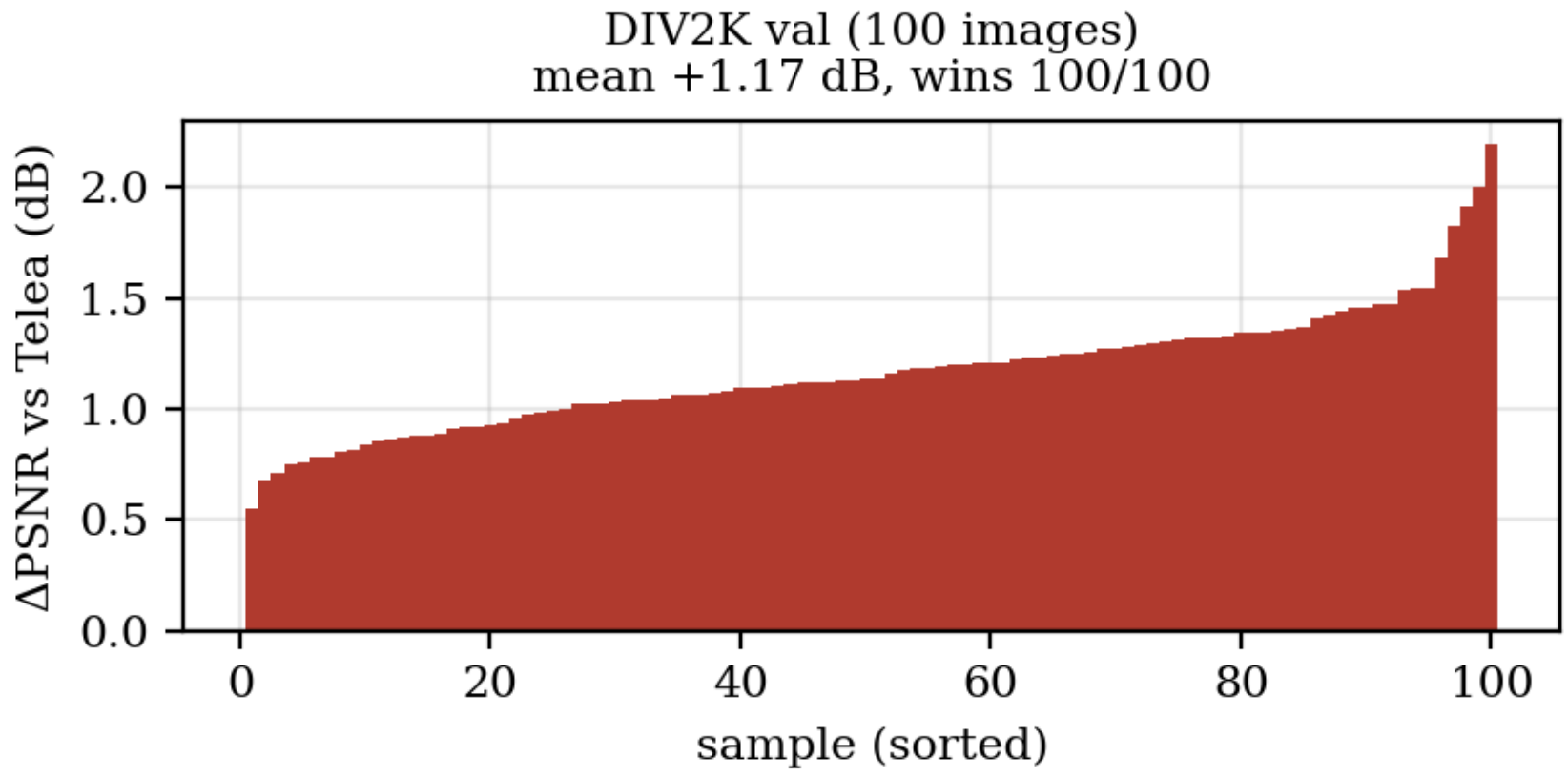

**Figure 8  Per-sample PSNR gain of v0.2 over Telea on DIV2K cracks (100 images, sorted). All 100 samples improve; mean +1.17 dB.**

### 4.8 Text occlusion

Text strokes are, geometrically, structural slits of a few pixels width, the regime in which the model is designed to excel. The text masks are generated programmatically from 2–4 random strings with stroke width 3–6 px and a mask ratio of about 5%. v0.2 beats Telea and NS-OpenCV on all 40 samples, gaining +1.61 dB over Telea on BSDS and +1.91 dB on the paintings, and ranks first or second among PDE solvers: on the paintings it reaches 46.067 dB, 0.15 dB behind the fractional-Laplacian model and ahead of every other baseline including LaMa at 45.226 dB, and on BSDS it ties the leaders within 0.3 dB, scoring 38.544 against 38.834 for LaMa and 38.688 for the fractional model. The absolute scores are high because the damaged area is small, but the ordering matches the crack experiments and supports the structural-slit positioning. Figure 9 shows a representative example: every method removes the text, but only the transport-type solvers restore the underlying texture without visible blur: the magnified insets of the marked text band show that CDD leaves a smeared band along the strokes, while double-CDD and GL-SAV introduce visible artifacts. Figure 10 summarizes both datasets.

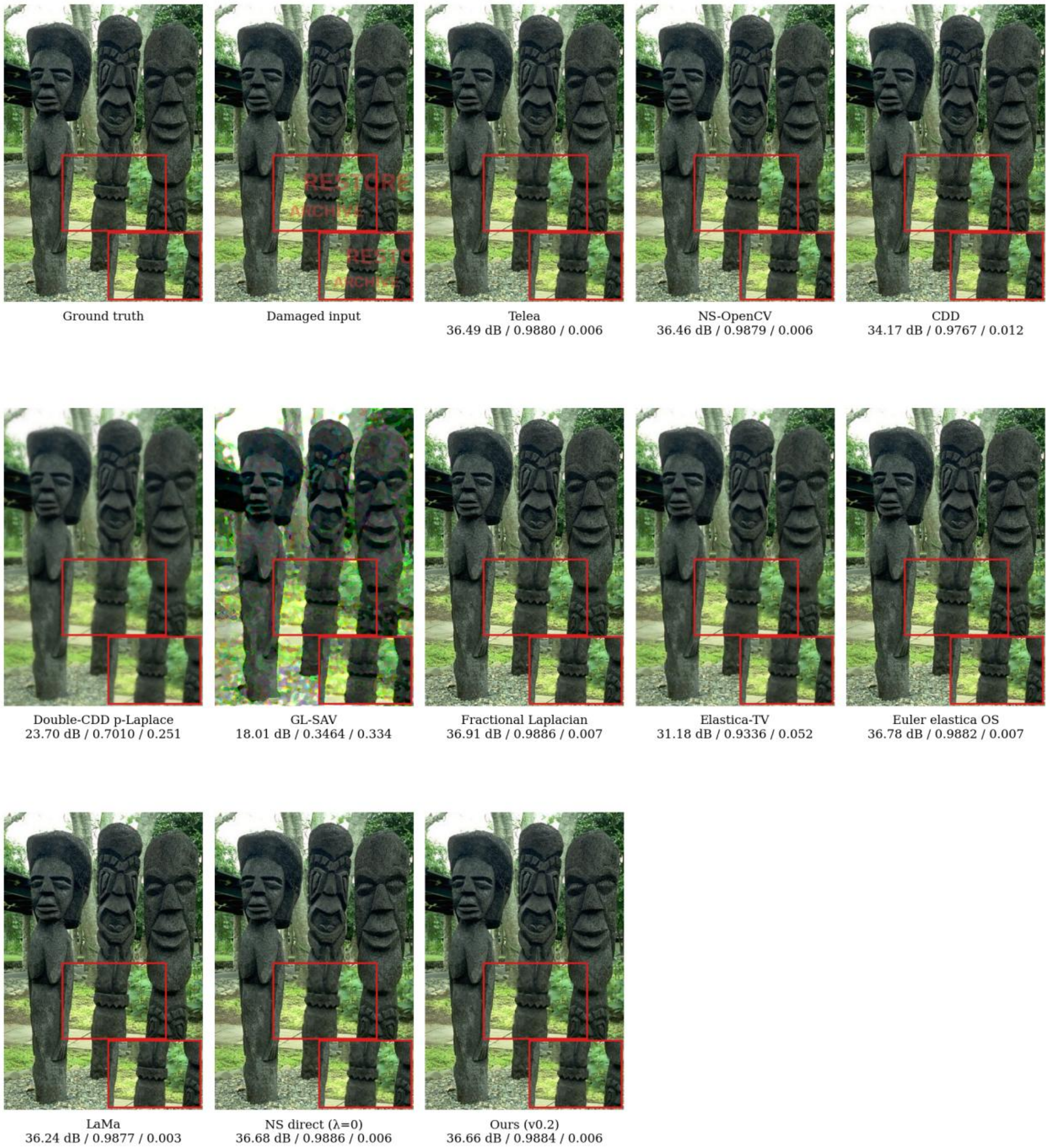

**Figure 9 Text-occlusion inpainting on a BSDS500 image (mask ratio 2.4%). Numbers beneath each method: PSNR (dB) / SSIM / LPIPS on this sample. The red rectangle marks the region magnified in the corner inset of every panel, covering the text band across the central totem.**

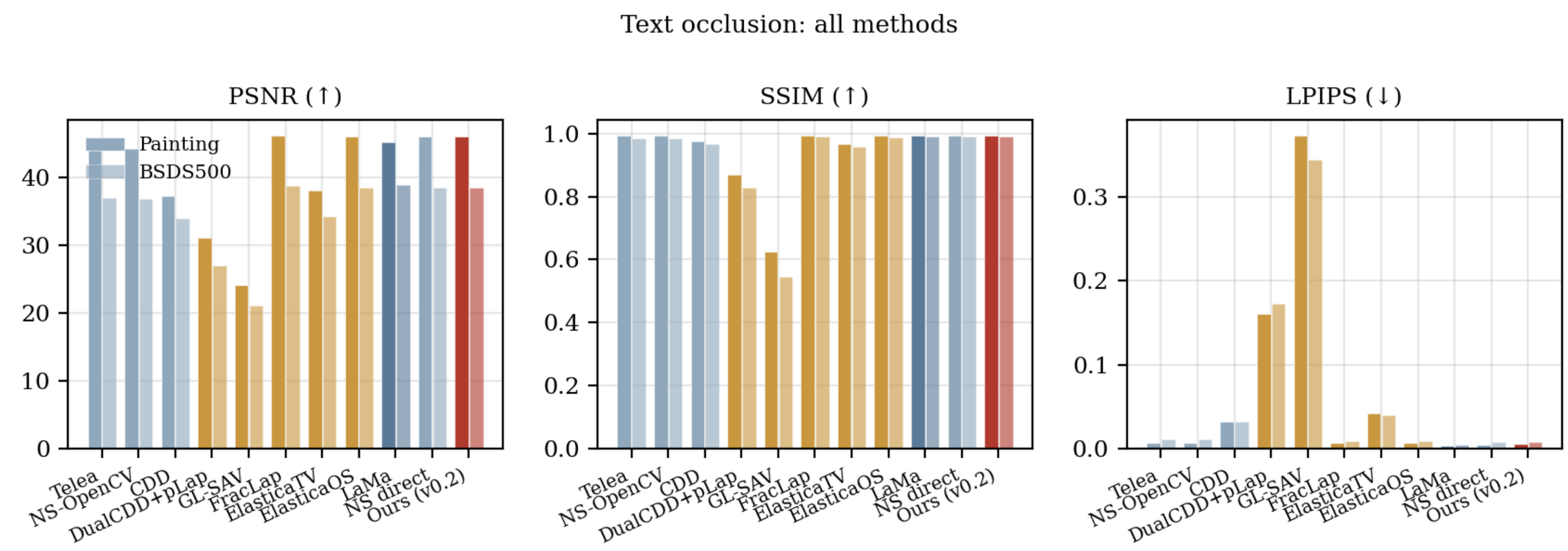

**Figure 10 Text occlusion (20 BSDS + 20 painting images), all twelve methods: mean PSNR and LPIPS. v0.2 ranks first or second among PDE solvers on both datasets and beats LaMa on paintings by +0.84 dB.**

## 4.9 Dual-direction stripes: a cautionary ablation

A natural refinement of stripe decomposition alternates two stripe orientations, 0° and 90° with four passes in total, to suppress one-directional artifacts. The ablation in Figure 11, run on 40 BSDS block samples with the same seeds as Section 4.5, gives a counter-intuitive result: dual direction improves the Telea-based variant substantially, by +0.28 dB, yet leaves the proposed model essentially unchanged at +0.04 dB, and the dual Telea variant overtakes the dual NS variant on PSNR with a 5/40 win rate for the latter. The interpretation is mechanical: the forward correction injects anisotropic structural detail in each slit pass, and the two additional passes of the second orientation partially diffuse that detail away. More passes do not help once the solver itself is strong. Model v0.2 therefore retains the single-direction configuration, and we recommend dual direction only when wrapping simple base fills.

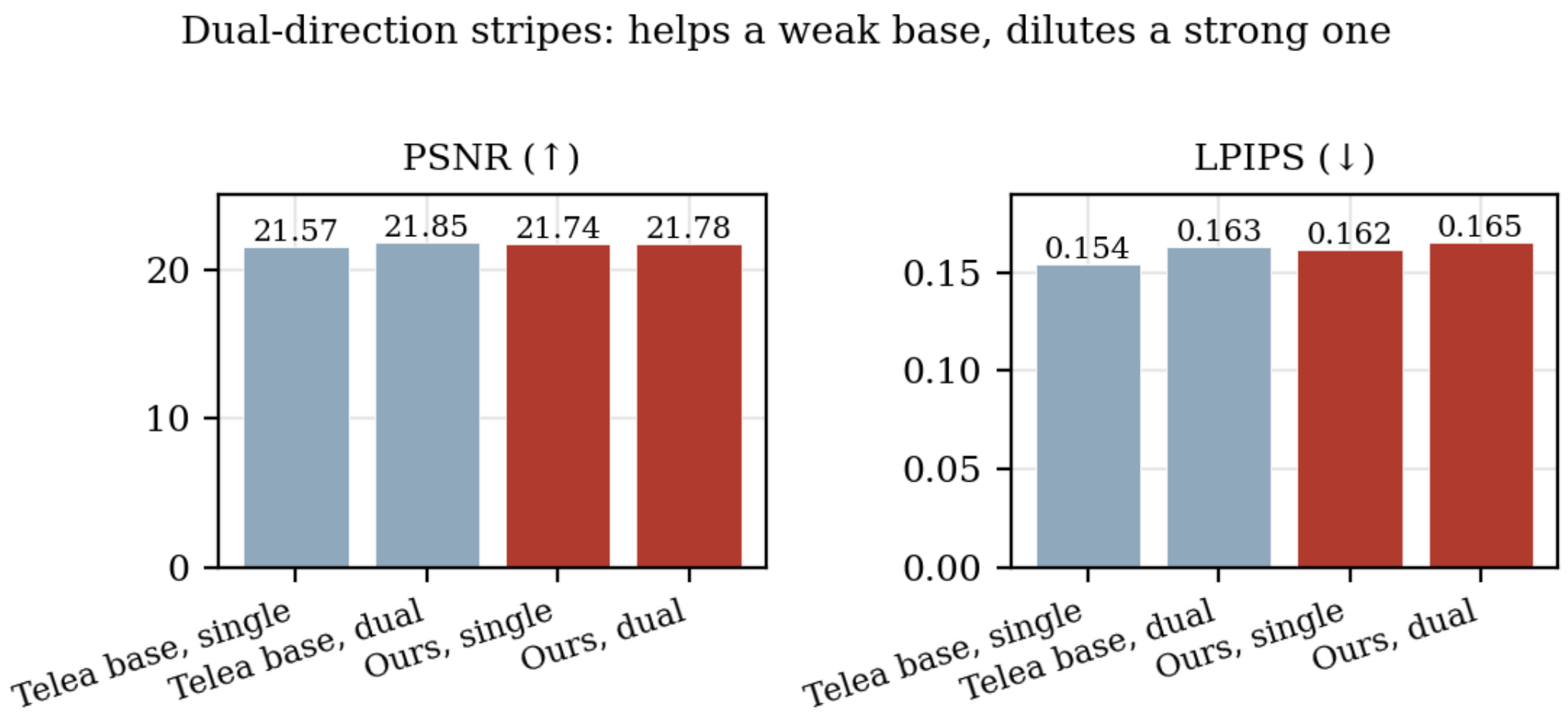

**Figure 11 Single- vs dual-direction stripes (BSDS blocks, 40 samples): mean PSNR and LPIPS. Dual direction helps the weak Telea base fill (+0.28 dB) but not the strong solver (+0.04 dB, per-sample win rate 5/40 against it).**

### 4.10 Diffusion-based reference: RePaint on a 256 × 256 subset

To position the proposed model against the diffusion family we evaluated RePaint, using the official 256 × 256 unconditional checkpoint with a 250-step resampling schedule, on a 30-sample subset consisting of 12 cracks, 12 blocks and 6 text occlusions drawn from all three datasets, with each image isotropically resized and reflect-padded to 256 × 256 and all other methods re-run on the same inputs. At about 540 s per image on our GPU the full 324-sample benchmark is impractical for RePaint; the subset keeps the mask generators and seeds of the main experiments.

The result, collected in Table 10, is unambiguous. As an unconditional generative model, RePaint hallucinates semantically plausible content rather than reconstructing the ground truth: the proposed model beats it in PSNR on all 30 samples by +3.47 dB on average, and even Telea outperforms it on every scenario. On LPIPS, RePaint wins 19 of 30 samples against v0.2 yet never reaches LaMa, and it costs about 80 times the runtime of the proposed solver, 535–539 s against 6.5–6.7 s per image. For fidelity-critical, auditable restoration of structural damage, an unconditioned diffusion prior is neither accurate nor economical. The two families answer different questions, and a well-posed physics prior is the right tool for the regime studied here.

**Table 10 RePaint vs selected methods on the 30-sample 256 × 256 subset (12 cracks, 12 blocks, 6 text occlusions)**

| Scenario / Method | PSNR↑ | SSIM↑ | LPIPS↓ | time (s) |
|---|---|---|---|---|
| Cracks / Fractional Laplacian | **26.666** | **0.8914** | 0.1077 | 0.1 |
| Cracks / v0.2 (ours) | <u>26.371</u> | <u>0.8880</u> | <u>0.0927</u> | 6.7 |
| Cracks / LaMa | 25.916 | 0.8879 | **0.0362** | 0.1 |
| Cracks / Telea | 25.200 | 0.8597 | 0.1135 | <0.1 |
| Cracks / RePaint | 22.575 | 0.8129 | 0.1032 | 538.7 |
| Blocks / Euler elastica OS | **23.746** | **0.8365** | 0.1792 | 0.2 |
| Blocks / v0.2 (ours) | <u>23.376</u> | <u>0.8342</u> | 0.1662 | 6.5 |
| Blocks / Telea | 23.269 | 0.8336 | 0.1752 | <0.1 |
| Blocks / LaMa | 22.557 | 0.8295 | **0.0608** | <0.1 |
| Blocks / RePaint | 21.113 | 0.7842 | <u>0.1085</u> | 535.2 |
| Text / Fractional Laplacian | **32.547** | **0.9641** | 0.0341 | 0.1 |
| Text / v0.2 (ours) | <u>32.272</u> | <u>0.9638</u> | <u>0.0279</u> | 6.5 |
| Text / LaMa | 31.827 | 0.9624 | **0.0136** | <0.1 |
| Text / Telea | 30.772 | 0.9513 | 0.0358 | <0.1 |
| Text / RePaint | 27.038 | 0.9211 | 0.0478 | 533.8 |

Summary of experimental findings. On structure-dominated damage such as cracks, scratches and text strokes, model v0.2 beats every classical baseline on every sample across three independent datasets and matches the strongest recent PDE solvers within 0.4 dB, while attaining the best LPIPS of the entire PDE family. On large blocks it delivers PSNR on par with the best classical baseline and the best perceptual quality among PDE methods, through a stripe decomposition whose gains are base-agnostic and additive with the forward correction; learning-based methods such as LaMa and RePaint dominate that regime. The sign of the correction term is decisive, confirming Theorem 1 experimentally, and the model is insensitive to λ over one order of magnitude. Two negative results, the failure of cheap feature gating in Section 4.5 and of dual-direction stripes in Section 4.9, delimit the design space and guard the claims against overfitting to a benchmark.

## 5. Discussion

### 5.1 Transport wins on structural slits

The experiments support a coherent physical picture. The NS solver is a transport device: it advects vorticity along isophotes, and it is at its best when the missing region is thin in the direction normal to the local structure, so that information only has to travel a short distance along a well-defined direction. Cracks, scratches and text strokes have exactly this geometry, and on all three the proposed model leads every classical baseline on every dataset, with per-sample win rates of 42/42 on painting cracks, 199/200 on DIV2K cracks and 40/40 on text occlusion. The forward fourth-order correction reinforces this regime: its gate activates on strong vorticity gradients, which concentrate along structural edges with 56% activation on crack masks, and the correction sharpens the transport front where it matters. The theory of Section 3 explains why this regime needs no tuning: the forward model is uniformly parabolic with an energy estimate, so the correction helps uniformly rather than conditionally, and the empirical insensitivity to λ over one order of magnitude is the experimental shadow of that theorem.

### 5.2 Blocks reward perception, not fidelity

On large missing blocks the picture is more nuanced and deserves a precise statement. With stripe decomposition the model achieves the best SSIM and LPIPS on the painting benchmark, while its PSNR is statistically tied with the best classical baseline, 0.11 dB behind NS-OpenCV and within the per-sample standard deviation; on BSDS blocks it leads all classical baselines on PSNR as well, by +0.45 dB over the direct classical solver and +0.34 dB over Telea. Two observations reconcile these facts. PSNR penalizes the unavoidable texture mismatch inside a smooth base fill, while SSIM and LPIPS reward the structural continuity that stripe decomposition actually restores, so the metric split reflects what the method does rather than noise. The exemplar-based Criminisi baseline inverts the split, ranking worst in PSNR and best in LPIPS on blocks, which shows that no classical method dominates block damage on all criteria and that claims on this damage type should be made per metric. We position the model accordingly: on block damage it is the perceptual-quality leader among classical methods, not a semantic synthesizer. Hallucination-grade completion of large semantic holes remains the territory of learning-based methods, to which our model is complementary in the sense of Section 6.

### 5.3 Cheap gating has no signal

A natural engineering question is whether the per-sample benefit of the higher-order correction can be predicted from cheap image features, so that the correction could be switched off where it does not help. We tested this with a small set of structural descriptors, including edge density, isophote coherence and gate activation rate, under a leave-one-out protocol. The answer is negative: no descriptor correlates with the per-sample gain beyond $|r| = 0.25$, and the best leave-one-out gating rule gains at most +0.08 dB over always-on, within noise and far below what the added complexity would require. The well-posedness theory explains the failure. The backward correction is conditionally useful in exactly the way gating assumes, helpful here and harmful there, and it is also ill-posed by Theorem 1; the forward correction is uniformly parabolic and energy-stable by Theorem 2, hence uniformly useful, and there is essentially no per-sample signal for a gate to exploit. Gating, in this setting, is the right question for the wrong sign.

### 5.4 Dual stripes dilute strong solvers

Stripe decomposition raises an obvious extension: if one family of parallel slits helps, two interleaved orthogonal families should help more, by giving every interior pixel structural support from two directions. The extension works for a weak base solver, where wrapping Telea with dual-direction stripes adds +0.28 dB over the single-direction variant, but it fails for the strong one. Wrapping the proposed model, the second family adds only +0.04 dB on average and loses the per-sample comparison with 5 wins in 40 samples, because it erodes the structure that the first pass had already reconstructed correctly. Two mechanisms explain the asymmetry. A strong solver leaves little headroom inside a slit, so the second family has little to recover and mostly re-damages: each additional pass is itself an inpainting problem with its own error budget, and for a near-ceiling solver that budget is spent rather than earned. The second family also imposes its own directional prior on content whose true structure is rarely aligned with both directions, and the resulting cross-hatch oversmoothing is what LPIPS detects. The practical rule is therefore solver-dependent: dual-direction stripes are a cheap booster for weak base fills and an anti-pattern for strong ones. Stripe decomposition is not a scheme in which more passes bring more gain, and we report the negative variant so that the design space is delimited honestly.

### 5.5 Three limitations bound the claims

Three limitations bound the claims. The correction gate is, by design, inactive where vorticity gradients are weak, with 7.3% activation inside smooth block interiors, so inside large blocks the model reduces to its second-order skeleton plus whatever structure the stripes inject; a spatially adaptive or learned gate could remedy this, but Section 5.3 shows that cheap variants do not work. All comparisons are within training-free classical methods, where the model is competitive at 4–16 s per image on a consumer GPU; against diffusion- or transformer-based inpainting the comparison is not on equal terms of compute or data, and we defer it to the hybrid-prior perspective of Section 6. The painting benchmark is domain-specific even though it is evaluated in full on 42 images without cherry-picking, and the BSDS block evaluation uses 40 samples; the per-sample win rates mitigate but do not eliminate the sample-size caveat. None of these limitations affects the theoretical results, which are asymptotic and parameter-independent.

## 6. Conclusion

This paper has carried the transport-type inpainting paradigm of Bertalmío, Bertozzi and Sapiro into the higher-order regime that diffusion-type models have occupied for two decades, and it shows that the move is safe only under explicit analytical conditions. The naive fourth-order correction, with the sign suggested by a regularization argument, is a Perona–Malik-type backward diffusion in the vorticity and is Hadamard ill-posed; the forward sign is uniformly parabolic and energy-stable, and the order–dissipation principle, stating that odd orders disperse and even orders dissipate, singles out the fourth order as the lowest admissible correction. Order selection, in this framework, is a theorem rather than a hyperparameter.

On the algorithmic side, stripe decomposition closes the remaining gap between the model's strength on thin, structure-dominated damage and the practically important case of large missing blocks, by converting a block into a sequence of slits in which the solver operates near its ceiling. The resulting training-free model beats every classical baseline on every sample of structure-dominated damage across three datasets, with 42 wins in 42 samples on painting cracks, 100 wins in 100 samples on DIV2K cracks and 40 wins in 40 samples on text occlusion, and it matches the strongest recent PDE solvers within 0.4 dB while leading the entire PDE family in perceptual quality as measured by LPIPS, and runs in 4–16 s per image on a consumer GPU. The two negative results belong to this conclusion as well: the forward correction needs no gating, and the stripe device needs no second direction, so the design space around the model is now mapped, and future users are protected from two plausible but fruitless modifications.

Three directions follow from these results. The most consequential is the use of the model as a physics prior in hybrid learning frameworks: the vorticity transport equation and the forward fourth-order correction are differentiable, interpretable and provably stable, and they can serve as a structure-preserving refinement stage on top of learning-based coarse completions, combining semantic synthesis with the auditable fidelity that heritage restoration demands. The gate-silence phenomenon suggests spatially adaptive or learned gates; our negative result rules out cheap descriptors, but a learned gate trained on solver-internal quantities rather than image features remains open. The order–dissipation principle invites a systematic treatment of sixth- and higher even-order corrections and of the three-dimensional and spatio-temporal settings of video inpainting, where the transport structure of the model may matter even more than in two dimensions.

A model's order, sign and gate should be derived before they are tuned, and negative results deserve the same care as positive ones; this paper has followed both rules. The code, masks and evaluation scripts are available, and every number in the paper can be reproduced end to end.

**Declaration of generative AI and AI-assisted technologies in the writing process**

During the preparation of this work the authors used Kimi (Moonshot AI) in order to assist with writing and language polishing. After using this tool, the authors reviewed and edited the content as needed and take full responsibility for the content of the publication.

## CRediT authorship contribution statement

Yiwen Dou: Conceptualization, Funding acquisition, Project administration, Supervision, Writing – original draft, Writing – review & editing. Yiqiang Ding: Methodology, Investigation, Formal analysis, Data curation, Visualization. Zhongyun Bao: Investigation, Validation, Resources, Writing – review & editing. Senyan Zhao: Software, Formal analysis, Data curation, Visualization, Writing – review & editing.

## Declaration of competing interest

The authors declare that they have no known competing financial interests or personal relationships that could have appeared to influence the work reported in this paper.

### Acknowledgements

This work was supported in part by the University-Level Development Projects of Anhui Polytechnic University (HX-202112008, HX-202412103) and the Synergetic Innovation Program of Anhui Provincial Universities (GXXT-2019-007).

### Data availability

The code, masks and evaluation scripts, together with the classical-painting benchmark used in this study, will be made publicly available upon publication.